# EVALUATING SELF-SUPERVISED SPEECH MODELS ON A TAIWANESE HOKKIEN CORPUS

*Yi-Hui Chou*[1†], *Kalvin Chang*[1†], *Meng-Ju Wu*[‡], *Winston Ou*[2‡], *Alice Wen-Hsin Bi*[3‡], *Carol Yang*[‡],
*Bryan Y. Chen*[4‡], *Rong-Wei Pai*[5‡], *Po-Yen Yeh*[6‡], *Jo-Peng Chiang*[7‡], *Iu-Tshian Phoann*, *Winnie Chang*[1],
*Chenxuan Cui*[1], *Noel Chen*[1], *Jiatong Shi*[1]

[1]Language Technologies Institute, Carnegie Mellon University
[2]Scripps College
[3]University of Maryland
[4]Swarthmore College
[5]National Taiwan Normal University
[6]China Medical University, Taiwan
[7]National Taiwan University

## ABSTRACT

Taiwanese Hokkien is declining in use and status due to a language shift towards Mandarin in Taiwan. This is partly why it is a low resource language in NLP and speech research today. To ensure that the state of the art in speech processing does not leave Taiwanese Hokkien behind, we contribute a 1.5-hour dataset of Taiwanese Hokkien to ML-SUPERB's hidden set. Evaluating ML-SUPERB's suite of self-supervised learning (SSL) speech representations on our dataset, we find that model size does not consistently determine performance. In fact, certain smaller models outperform larger ones. Furthermore, linguistic alignment between pretraining data and the target language plays a crucial role.

## 1. INTRODUCTION

This paper presents a diverse natural speech dataset of Taiwanese dramas, contributing a 1.5-hour dataset of Taiwanese Hokkien to ML-SUPERB's hidden set. We validate the dataset on the ASR task, finetuning and evaluating ML-SUPERB's suite of self-supervised learning (SSL) speech representations on our dataset. We find that larger model sizes do not consistently result in better performance. While some larger models outperform their base counterparts, model success is more heavily influenced by the quality and linguistic alignment of the training data. Furthermore, we analyze the performance of a variety of self-supervised speech representations on our dataset, and reveal phonetic similarities, tone, audio quality issues, and the omission of particles as sources of prediction errors.

†, ‡Equal contribution

## 2. TAIWANESE HOKKIEN

Taiwanese (or Taiwanese Hokkien, tâi-gí/tâi-gú, Taiwanese Southern Min) is a unique variant of the Hokkien language spoken by 18 million Taiwanese people to varying degrees of fluency [1], accounting for nearly 80% of the population.

### 2.1. Linguistic properties

Taiwanese and Mandarin are both Sinitic (Chinese) languages, exhibiting similar morphological and syntactic properties. Words in Sinitic languages are made of a single character (or one syllable), two (in compounds) [2], or more. Both Taiwanese and Mandarin employ SVO word order and use sentence-final particles [3]. However, the two are mutually unintelligible due to vastly different pronunciations. As a Min variety, Taiwanese did not undergo a Middle Chinese phase, unlike Mandarin [4]. Thus Taiwanese is even more dissimilar from Mandarin than other descendants of Middle Chinese, such as Yue. In addition, the 7 tones of Taiwanese interact in a complex web of tone sandhi that applies to all syllables whenever they do not occur in the final position of specific syntactic units [5].

### 2.2. Orthography

Taiwanese is not widely written down by speakers of any age group today. Instead, written correspondence largely occurs in Mandarin. When the situation requires written Taiwanese, methods of writing include, but are not limited to, the following:

1. **Romanization**: Two relatively prominent romanization systems are pe̍h-oē-jī (invented by Western missionaries in the 19th century) and tâi-lô (derived from the former and promoted by the Ministry of Education in Taiwan). Although the romanization systems were designed to increase literacy in Taiwanese, they are not widely used among all speakers, many of whom have not learned romanization [6].

2. **Taiwanese Han**: The Ministry of Education released a set of Han characters for use in Taiwanese [7], which can be used in tandem with romanization (when no Han character exists) to form hàn-lô. However, in more informal settings, speakers often substitute Taiwanese characters with phonetically similar but semantically different characters commonly used in Mandarin (台語火星文) [6]. The usage of these is *ad hoc* and often inconsistent among different speakers.

Below is a sentence from our corpus written in Taiwanese Han and tâi-lô romanization, with the official Mandarin subtitles and our own English gloss below. Mandarin speakers may have trouble understanding either method of Taiwanese orthography. On the third line, we include the result of our preprocessing (§5.1). Note that our dataset is in tâi-lô romanization, not Taiwanese Han.

1. 我　想欲　　看　　伊　未來　會　變做　　啥物　　　款
   guá siūnn-beh khuànn i   bī-lâi ē   piàn-tsó siánn-mih khuán
   g ua2 si iunn7 b eh4 kh uann3 i1 b i7 l ai5 e7 p ian3 ts o2 si iann2 m ih4 kh uan2

   "I want to see what they will be like in the future"

   「我想看他未來會成為什麼樣子」

Both tâi-lô and pe̍h-oē-jī use diacritics to mark tones, and only the base or citation (pre-tone sandhi) tones are indicated in text. For example, the base tones of "tán--tsit-ē" ("wait a minute") are, in order, 2, 8, and 7. Single dashes are used to connect morphemes that belong to the same word (e.g. tsit-ē). Unlike Han characters (in either Mandarin or Taiwanese), romanization comes with word segmentation by default. Double dashes indicate that the syllable or word to their immediate right become the neutral tone.

### 2.3. Sociohistorical context

This variety was born out of contact between the tsuân-tsiu (Quanzhou) and tsiang-tsiu (Zhangzhou) Hokkien varieties brought by Han settlers to Taiwan. During Japanese colonization (1895-1945), Taiwanese acquired many loanwords from Japanese, unlike other varieties of Hokkien. During martial law, the KMT regime banned all non-Mandarin languages (Formosan languages, Taiwanese Hakka, and Taiwanese Hokkien), leading to a language shift in favor of Mandarin that persists today [6]. Unsurprisingly, Mandarin has influenced Taiwanese pronunciation and word choice [6]. A more serious consequence of the shift is that Taiwanese and Mandarin now exist in a diglossic situation for many speakers: the former is used for conversation; the latter, for administration and education. However, only 7.4% of those age 6 - 14 spoke Taiwanese as their main language in 2020 [1]. This language ideology has carried over to NLP and speech, where there is a dearth of Taiwanese text and speech corpora compared to Mandarin. Recent efforts have included Taiwanese in NLP and speech [8, 9, 10, 11], however, and we aim to contribute to this momentum with our dataset.

## 3. DATASET

Our dataset consists of Taiwanese soap opera speech, gold Mandarin subtitles, and Taiwanese annotations. Soap opera speech presents a challenging testing environment for speech models because it involves diverse prosodic features, among other paralinguistic cues (such as crying and laughing).

### 3.1. Statistics

The speech data was converted into a sampling frequency of 16kHz. As the specific number of speakers was not explicitly provided, we consider each utterance as being spoken by a different speaker.

Table 1 provides statistics related to our dataset of Taiwanese dramas. The table encompasses crucial information for each data split, including the total duration of the speech data in minutes, the count of utterances, the average length of utterances measured in seconds, the average number of syllables per utterance, and the total number of dramas included in the dataset.

**Table 1**. Dataset Statistics for Taiwanese Dramas in Train (10 Minutes and 1 Hour), Development, and Test Sets.

| Split | #mins. | #utts. | Avg. length of utts (secs) | #syllables | #dramas |
|---|---|---|---|---|---|
| Train (10 mins) | 10.16 | 308 | 1.98±0.86 | 13.41±5.32 | 4 |
| Train (1 hour) | 60.34 | 1708 | 2.12±0.80 | 15.15±5.79 | 17 |
| Dev | 10.07 | 320 | 1.88±0.69 | 15.31±6.35 | 4 |
| Test | 10.02 | 275 | 2.16±0.91 | 15.60±6.02 | 5 |
| Total | 90.60 | 2611 | 2.08±0.81 | 15.01±5.86 | 30 |

### 3.2. Design decisions

We transcribe with tâi-lô romanization instead of pe̍h-oē-jī and follow the orthographic conventions of Taiwan's Ministry of Education [12]. During transcription, we use diacritics for tones (as opposed to number tones) since Taiwanese keyboards output diacritics (e.g. tâi-gú instead of tai5-gu2). In line with [12], we mark the base (or citation) tones for all syllables, which means that speech models must learn to associate the sandhi tones in the speech with the base tone. We preserve dialectal variation when possible. For instance, we transcribe 今仔日 as kin-á-lit, as opposed to standardizing it to kin-á-jit, when the speaker uses [l] instead of [d͡z] or [d͡ʑ].

Because Taiwanese particles (e.g. --lah) function as discourse markers [3], we include them in the transcriptions. We mark particles even when they occur mid-utterance. We also mark syllable contractions (合音) when used—which are abundant in the audio according to the annotators—as opposed to expanding them (e.g. buǎ-kín over bô-iàu-kín). We train on Taiwanese romanization as opposed to Han characters, because the latter does not reveal the pronunciation of the character, while romanization enables the model to exploit the high degree of homophony in Sinitic languages and share information across different characters with the same phones involved.

## 4. DATASET CURATION

### 4.1. Source

We crawled a dataset of Taiwanese dramas from YouTube using the YouTube-ASR-Crawler tool [1] for academic purposes only. We do not own or distribute this data, which carries the standard YouTube license. The dataset primarily consists of dialogue in Taiwanese Hokkien, with Mandarin subtitles provided. Our choice of using Taiwanese dramas as the data source was mainly driven by the availability of natural speech on such a large scale and the diverse range of conversational scenarios they offer. [2]

To ensure the quality of the data, we sampled sentences without recurring themes (such as character names and pronouns) and excluded clips involving code-switching overlapping speech. This resulted in 2,613 sentences from 53 episodes from 31 dramas. We made sure that each drama was exclusively present in only one of the data subsets. Given that dramas do not contain speaker identities (actors and actresses), we considered each segment of speech to be spoken by a different speaker. [3]

To extract audio segments, we utilized the starting and ending times of subtitles, which occasionally led to missing words at the beginning and end of the speech segments. To address this issue, we performed a manual alignment procedure for the train, dev, and test sets, except for the train (1 hour) subset due to time constraints. This resulted in an average utterance length of $2.0 \pm 0.8$ seconds, with an average of $15.0 \pm 0.9$ syllables per utterance [4].

The dramas in our dataset present various acoustic effects, including background music, sound effects, room acoustics, occasional background chatter, and overlapping speech. While these properties can make speech data more challenging to work with, they also make it more representative of real-world communication scenarios. To address the issue of background music, which is prevalent in dramas, we employed a music source separation model [5] [15] to remove it from the audio.

### 4.2. Annotation

The Taiwanese labels for our dataset were generated by the authors who were skilled in Taiwanese transcription. The annotators (n=10) consist of Taiwanese and Taiwanese American, native and L2 speakers of Taiwanese Hokkien (6 native, 4 L2), with an average of $22.9 \pm 18.8$ years of experience speaking or studying the language. They and their parents represent a diverse set of cities (and thus dialects): Taipei, Kaohsiung, Tainan, Taichung, Hsinchu, Chiayi, Yunlin, and Pingtung. During the first pass, annotators generated labels, with Mandarin subtitles and machine translated Taiwanese as guides. During the second pass, annotators helped each other resolve specific words or phrases they could not clearly hear.

#### 4.2.1. Machine translation-assisted annotation

To reduce the amount of typing transcribers have to perform, we provide transcribers with machine translated Taiwanese generated from the Mandarin subtitles. We use Fairseq [16] to train a standard encoder-decoder Transformer model for 80 epochs using Sentencepiece [17] BPE tokenization, an Adam optimizer [18], a dropout of 0.3, and a beam size of 5 for decoding. The model is trained on the iCorpus dataset [11], a Mandarin-Taiwanese parallel corpus of 83,544 sentences collected from 3266 news articles. We use Fairseq's `transformer_iwslt_de_en` model architecture which has 6 layers and 4 attention heads for the encoder and decoder each. The MT (machine translation) module achieves a BLEU score [19] of 78.41 and a COMET score [20] of 0.7694 on the iCorpus [11] test set, which is unsurprising given the similarity in morphology and syntax between the two languages. However, because our corpus of dramas presents a domain shift for the model trained on news, the output is not always accurate, which is why manual transcription is still needed.

Many of the errors involve incorrect word segmentation. For instance, instead of predicting "guá siūnn-beh khuànn i bī-lâi ē", MT predicts "guá-siūnn-khuànn-i-bī-lâi-huē." In addition, the word choice of the MT output may differ from the word chosen by the speaker (e.g. "bô-kuan-hē" instead of "buǎ-kín", which both mean "no problem"). While this technically is not an error of MT, it still reflects the need for human annotation.

In short, significant revision for most of the machine translations is needed for accurate annotation. We instruct all annotators to listen to each audio clip to which they are assigned and to transcribe them in Taiwanese romanization. The MT output thus serves only as a reference.

---

[1] https://github.com/Chung-I/youtube-asr-crawler

[2] We have explored various sources of Taiwanese corpora, but they either have proprietary restrictions [9], are too limited in size (Suisiann [13], CommonVoice-Hk [14], and MOE dict's sample sentences), or are not accessible without cost (Taiwanese Across Taiwan [8] requires a fee of US $1,500).

[3] What is clear is that the dataset contains speakers of both genders, various ages (young children all the way to elderly individuals), and many dialects beyond the Kaohsiung standard dialect used by the Ministry of Education.

[4] Standard deviation is reported after ±.

[5] github.com/facebookresearch/demucs

## 5. EXPERIMENTS

In this paper, we conduct a comprehensive investigation into the efficacy of speech recognition models on the Taiwanese dataset we collected using the ESPnet toolkit [21].[6]. We report the character error rate (CER) and syllable error rate (SER) across various experimental setups. When CER is calculated, character refers to one Latin character (e.g. honnh has 5 characters), not one Han character or syllable.

### 5.1. Preprocessing

To ensure the model learns more effectively, we apply preprocessing techniques. During training, we convert from diacritics for tones to number tones using a Taiwanese text processing toolkit[7]. We then remove dashes from words, which means the WER reported by ESPnet is actually closer to syllable error rate (SER). We also find that character-level tokenization where every Latin character is one token is too fine grained. Thus we tokenize syllables into initials and finals using a custom tokenizer [8], in line with how historical Sinitic phonology analyzes syllables. This further increases the information sharing across different words that may have the same final but different initial (e.g. luảh and puảh) and vice versa. We reflect this coarser tokenization in ESPnet by switching the token type from character-level to word-level. Here is an example summarizing our preprocessing steps: khuànn--khílâi → khuann3--khi2-lai5 → khuann3 khi2 lai5 → kh uann3 kh i2 l ai5. Note that the model predicts initials and finals, not syllables. Because the predictions (which come in this preprocessed format) are often malformatted, we cannot detokenize the initial and final back into the original syllable. We therefore calculate the error rate treating initials or finals (e.g. "kh" or "uann3") as one token and refer to this as SER.

### 5.2. Model

We follow the standard training procedure and configuration outlined in the ML-SUPERB GitHub repository [9]. The ASR training process, as discussed in their paper [22], starts with a weighted sum of frozen self-supervised speech learning (SSL) representations, where the weights are learnable. SpecAugment is applied to the weighted sum of speech SSL representations. Subsequently, a convolutional downsampling layer is used to reduce the sequence of features by half, after which these hidden states are passed to a Transformer model. For optimization, we employ the Connectionist Temporal Classification (CTC) loss function with a dropout rate set at 0.1, and we utilize the Adam optimizer with a learning rate of 0.0001.

To explore the benefits of SSL representations, we incorporate twelve distinct SSL models suggested by ML-SUPERB, alongside the conventional fbank representation, into our investigation. By employing this diverse range of models, we aim to comprehensively understand their individual strengths and limitations to inform future work in our vein.

As Table 2 shows, the fbank representation exhibits limited effectiveness as anticipated, due to its simplistic nature. On the other hand, the integration of self-supervised learning (SSL) representations enhances performance significantly.

## 6. DISCUSSION

A noteworthy observation in our investigation is that superior performance is not solely dependent on model size. Surprisingly, both wav2vec2 and HuBERT base models outperform or have a similar performance to their larger counterparts, suggesting that model complexity may not be the sole determinant of success.

In fact, the impact of model size is context-dependent. Although `wav2vec2-base-23`, built on a larger and diverse multilingual dataset, underperforms on our specific dataset, HuBERT's larger variant (`hubert-large-cmn`) outperforms its base model (`hubert-base-cmn`). This discrepancy highlights the significance of the data on which larger models are trained, as exemplified by the success of `hubert-large-cmn`, which benefits from a larger and better-aligned language dataset.

Comparative evaluations among various model families reveal intriguing insights. The XLSR series, representing encompassing the self-supervised cross-lingual speech representation approach, outperforms both the wav2vec2 and HuBERT families, except when the models are trained on Mandarin. The robustness and adaptability of XLSR representations are indicative of their potential in diverse multilingual settings.

`mHuBERT-base`, pre-trained on a dataset comprising three languages and encompassing a substantial 13.5k hours of speech data, outperforms `HuBERT-base`, indicating that an increase in pre-training data does confer certain advantages. The ability of `mHuBERT-base` to leverage a more extensive and diverse dataset contributes to its improved performance compared to the HuBERT model trained on a smaller dataset. However, despite mHuBERT's superiority over HuBERT trained on 960 hours, it does not surpass `HuBERT-base-cmn`, a variant trained on a smaller dataset of 10k hours specifically in Mandarin. This finding suggests that merely increasing the quantity of pretraining data is not the sole determinant of better performance; the linguistic alignment between the pretraining data and the target language plays a crucial role. The phonological, morphological, and syntactic similarity between Mandarin and Taiwanese (§2.1) enable `HuBERT-base-cmn` to capture and transfer relevant represen-tations, enhancing its ASR performance on Taiwanese data. This highlights the importance of data quantity and linguistic alignment in speech representation learning.

---

[6]Configuration available at https://github.com/sophia1488/ML-SUPERB-on-TW-HK
[7]https://github.com/i3thuan5/tai5-uan5_gian5-gi2_kang1-ku7.
[8]https://github.com/wchang88/Tai-Lo-Tokenizer
[9]https://github.com/espnet/espnet/tree/master/egs2/ml_superb/asr1#monolingual-asr

**Table 2**. Character Error Rate (CER) and Syllable Error Rate (SER) of ASR using different upstream speech self-supervised learning (SSL) models on test set, trained with 10 mins and 1 hour.

|  | 10 mins | | | | 1 hour | | | |
|  | Dev | | Test | | Dev | | Test | |
| Model | CER | SER | CER | SER | CER | SER | CER | SER |
| --- | --- | --- | --- | --- | --- | --- | --- | --- |
| fbank | 69.5 | 84.7 | 69.1 | 84.5 | 54.1 | 77.3 | 54.8 | 78.5 |
| wav2vec2-base [23] | 41.5 | 58.7 | 41.7 | 59.5 | 32.9 | 47.7 | 32.2 | 47.6 |
| wav2vec2-large | 42.6 | 59.6 | 42.8 | 60.1 | 31.9 | 46.1 | 32.5 | 47.6 |
| robust-wav2vec2-large | 41.8 | 59.4 | 40.9 | 58.0 | 32.7 | 46.2 | 30.9 | 45.4 |
| wav2vec2-base-23 | 43.6 | 61.9 | 43.5 | 61.8 | 32.2 | 48.5 | 33.1 | 49.3 |
| wav2vec2-large-23 | 42.6 | 60.0 | 42.9 | 60.2 | 31.4 | 46.3 | 30.8 | 45.9 |
| XLSR-53 [24] | 39.7 | 57.7 | 38.9 | 57.1 | 30.3 | 43.8 | 30.0 | 44.1 |
| XLSR-128 [25] | 36.9 | 52.0 | 37.3 | 52.4 | 28.7 | 41.1 | 27.8 | 40.6 |
| HuBERT-base [26] | 41.0 | 57.2 | 41.2 | 58.2 | 32.9 | 47.5 | 33.0 | 48.0 |
| HuBERT-large | 43.2 | 60.3 | 41.4 | 58.4 | 34.8 | 49.9 | 33.6 | 48.4 |
| HuBERT-base-cmn | 35.6 | 51.3 | 36.5 | 52.1 | 26.6 | 39.0 | 25.8 | 38.6 |
| HuBERT-large-cmn | **31.4** | **46.5** | **31.6** | **45.9** | **23.5** | **35.0** | **22.7** | **33.7** |
| mHuBERT-base [27] | 39.6 | 57.1 | 40.3 | 58.0 | 32.9 | 47.4 | 31.8 | 46.9 |

## 7. ANALYSIS

We examine the performance of our best ASR model, `HuBERT-large-cmn`, on the test set of our Taiwanese dataset. Several key observations emerged from our evaluation of a subset of the model's predictions. First, the model tends to confuse phonetically similar sounds, such as predicting "tsi" [t͡ɕi] instead of "ji" [d͡zi] (both of which include affricates), "o" [o] instead of "oo" [ɔ] (both of which are back rounded vowels that have undergone a merger for many speakers of the Taipei dialect [28]), and "h" [ʔ] instead of "p" [p] (both of which are stops). We cannot exclude the possibility that such confusions are due to errors in pronunciation by the actors and actresses, which many of the annotators have reported. Second, we observed tone errors in the predictions, with some errors aligning with tone sandhi rules, such as the transformation of Tone 5 to 3 or 7, 1 to 7, 3 to 2, and interchange between tones 4 and 8. For instance, the model predicted 7 instead of the base tone 1 in one instance. While a syllable with base tone 1 that has undergone tone sandhi should technically be read as tone 7, if we want the model to output correct romanization, the base tone should be outputted. In another instance, the model predicted 1 instead of the base tone 7. Some predicted segments even lack any specified tone. Third, we encountered audio quality issues that led to errors in the transcriptions. Lastly, inconsistencies arose due to certain transcribers omitting particles, which subsequently impacted the model's ability to predict accurately, particularly concerning syllable contraction cases. These findings provide valuable insights into the strengths and limitations of the `HuBERT-large-cmn` model's performance on this dataset, highlighting areas for improvement in Taiwanese ASR.

To gain a deeper understanding of the impact of tone errors on the overall error rate, we conducted a manual exclusion of tones from the transcriptions and recalculated the Syllable Error Rate (SER), employing our own implementation of Levenshtein edit distance to calculate the SER instead of ESPnet's. This process resulted in a 16.6% relative error reduction of the SER. Moving forward, engineering efforts for Taiwanese ASR should focus on handling tone sandhi properly.

## 8. CONCLUSION AND FUTURE WORK

We contribute a dataset of 1.5 hours diverse and natural speech collected from Taiwanese dramas to ML-SUPERB's hidden set. We find that larger model sizes did not consistently lead to better performance. While certain larger models outperform their base counterparts, the key determinant of model success is primarily the quality and linguistic alignment of the training data. Our experiment provides key insights and challenges for automatic speech recognition of Taiwanese Hokkien. Specifically, we observe tone sandhi errors and confusing phonetically similar phones as sources of errors in the predictions. Audio quality issues also impacted the model's accuracy. Future research on Taiwanese ASR should prioritize improving tone accuracy and robustness against phonetic similarity and poor audio quality. This could involve a separate classifier for tone prediction that will learn tone sandhi—similar to how some TTS models have a separate model to predict prosodic attributes or suprasegmental features such as duration [29].

With the full several hundred hour dataset of dramas, we will attempt Taiwanese-to-Mandarin speech-to-text translation. The Taiwanese annotations in the 1.5 hour subset we discuss in this paper can provide auxiliary ASR supervision for this speech translation model [30].

Beyond our work, Taiwanese speech processing in general can encourage increased use of the language, reversing the language shift towards Mandarin and sparking a renaissance of this declining language.

## 9. ACKNOWLEDGEMENTS

This research is part of the Delta research computing project, which is supported by the National Science Foundation (award OCI 2005572), and the State of Illinois. Delta is a joint effort of the University of Illinois at Urbana-Champaign and its National Center for Supercomputing Applications.

We would like to express our utmost thanks to Soh-Eun (Ryan) Shim for helping us find transcribers, Jefferson Hsieh for helping with annotations, David R. Mortensen for answering our questions regarding dataset curation, and Lori Levin and Alexander Rudnicky for their guidance. We also appreciate the valuable feedback from the reviewers and the area chair.